\documentclass[a4paper, 10pt, conference]{mhs}      

\IEEEoverridecommandlockouts                              

\usepackage{graphics} 
\usepackage{epsfig} 
\usepackage{mathptmx} 
\usepackage{times} 
\usepackage{amsmath} 
\usepackage{amssymb}  
\usepackage{here}
\usepackage{threeparttable}
\usepackage{caption}

\title{\Large \bf
LLM-Based Human-Robot Collaboration Framework for Manipulation Tasks\\
}

\author{Haokun Liu$^1$, Yaonan Zhu$^1$$^{\ast}$, Kenji Kato$^2$, Izumi Kondo$^2$, Tadayoshi Aoyama$^1$, and Yasuhisa Hasegawa$^1$\\
1. Department of Micro-Nano Mechanical Science and Engineering, Nagoya University,\\
Nagoya, Aichi, 464-8603, Japan\\
2. National Center for Geriatrics and Gerontology,\\
Obu, Aichi, 474-8511, Japan\\\\
\thanks{$^{\ast}$Corresponding author email: zhu@robo.mein.nagoya-u.ac.jp}
\thanks{This work has been submitted to the IEEE for possible publication.
Copyright may be transferred without notice, after which this version may
no longer be accessible}}

\begin{document}
	
	\maketitle
	\thispagestyle{empty}
	\pagestyle{empty}

	\begin{abstract}
		This paper presents a novel approach to enhance autonomous robotic manipulation using the Large Language Model (LLM) for logical inference, converting high-level language commands into sequences of executable motion functions. The proposed system combines the advantage of LLM with YOLO-based environmental perception to enable robots to autonomously make reasonable decisions and task planning based on the given commands. Additionally, to address the potential inaccuracies or illogical actions arising from LLM, a combination of teleoperation and Dynamic Movement Primitives (DMP) is employed for action correction. This integration aims to improve the practicality and generalizability of the LLM-based human-robot collaboration system.
	\end{abstract}

	
	\section{Introduction}
    As robotics technology advances, the potential for robots to assist with domestic chores becomes increasingly promising. With the ability to understand and process natural language, these robots become more adaptable and flexible to accommodate a wide range of user instructions\cite{10161317}. However, the previous works with LLM-based control sometimes show a relatively low accuracy for high intelligence task decision-making\cite{10161317}. Our work introduces the idea of "LLM-Based task planning with human-robot collaboration", which is a novel approach to enhance human supervision in LLM-based autonomy. The contributions of this paper are summarized as follows: (1) Our LLM converts high-level language commands into sequences of executable motion functions, enabling adaptability to various user instructions. (2) Additionally, teleoperation and DMP are utilized for motion demonstration which allows the robot to learn from human guidance and potentially improves task feasibility and generalizability. (3) Furthermore, the robot is empowered with environmental perception through YOLO-based perception module for targeted tasks. The position of objects will be registered once recognized and update with the real-time position. Combining these elements, the proposed approach opens new possibilities for seamless human-robot collaboration in household tasks, making robots more practical and adaptable.

	\section{LLM-based human-robot collaboration framework}
\begin{figure}[t]
     
    	\centering
    	\includegraphics[width=\linewidth]{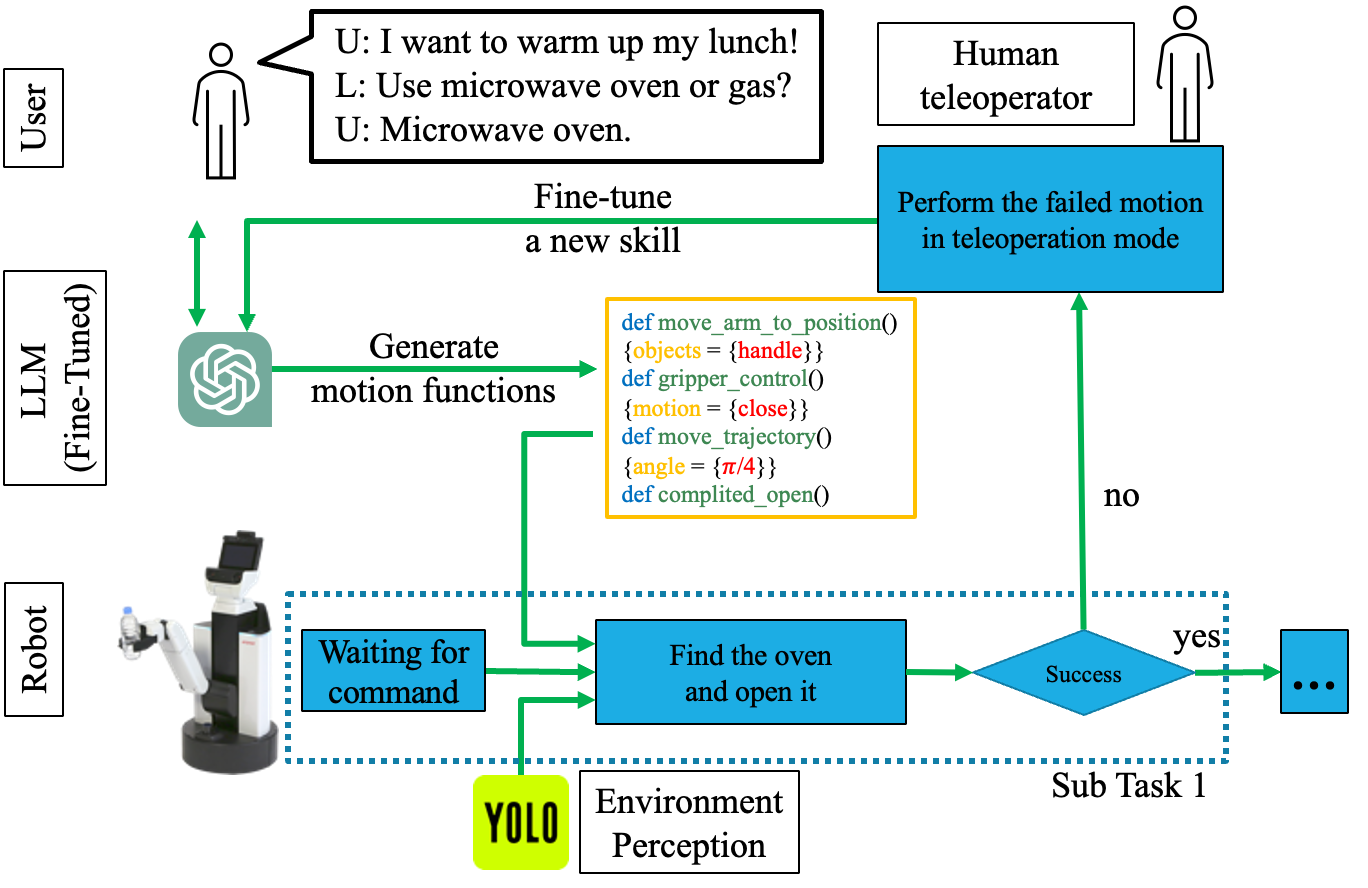}
    	\caption{ Framework of LLM-based task planning with enhanced HRC.  }
    	\label{fig:system}
     \vspace{-8mm}
    \end{figure}

    The system diagram is illustrated in Fig. \ref{fig:system}. The system consists of three main components: the user, LLM, and the robot, which forms an interactive loop. Additionally, we introduce a skilled teleoperator as an assistant to enhance the overall system's generalizability and feasibility.
    \subsection{LLM-based task planning}
    In our approach, we build our model based on LLM (GPT-2) and train it using a text corpus following previous work done by other researchers\cite{DBLP:journals/corr/abs-2305-07716}, enabling LLM to provide accurate function predictions in response to specific instructions. Subsequently, we integrate the perceived target position information and motion functions obtained from LLM into a prepared code template, enabling the robot to execute the corresponding tasks effectively. 
    \captionsetup[figure]{labelsep=period}
    To efficiently manage task execution, we adopt a hierarchical approach in our work-treating long-horizon tasks, short-horizon tasks, and motion functions as three layers. For long-horizon tasks which include motion functions of more than 10, we consider them first-layer tasks. In such cases, these tasks are separated into multiple short-horizon tasks through LLM. However, short-horizon tasks which involve less than 10 motion functions, are treated as the second layer task. When LLM receives commands from these second-layer tasks, it directly returns the functions necessary to accomplish the designated tasks. 
    
    On the other hand, the first-layer tasks can be separated into multiple short-horizon tasks through LLM. Subsequently, we process each short-horizon task following its specific procedure to divide them into motion functions as mentioned. Finally, motion functions are organized by following a planned task sequence to construct the complete long-horizon task. This hierarchical task handling allows for a more organized and effective execution of both short and long-horizon tasks, contributing to our system's overall efficiency and accuracy.
   
	
	\subsection{DMP-based task correction}
    To enhance the generalizability of LLM-based autonomy, We propose to integrate DMP-based task correction with human teleoperation-driven demonstrations. Dynamic Movement Primitives (DMP) is a generic approach for trajectory modeling in an attractor landscape based on differential dynamical systems \cite{7759554}. In this paper, we leverage our previously developed teleoperation system\cite{nakanishi2020towards,10173494} which can intuitively control the robot motion through a VR device, and also utilize DMP to record trajectories obtained from manual teleoperation. These trajectories can then be reproduced to complement any deficiencies in the LLM-based autonomy, particularly in failed function sequence generation or function sequence impracticality.
    
    For instance, when we issue the command "catch the bowl," the default motion function for bowl grasping could be inadequate to complete the task. To address this issue, we switch to the DMP-based teleoperation mode and provide instructions for the desired action. The robot can then accurately reproduce the trajectory using DMP. This approach will be continually developed to manage a wider range of long-horizon tasks, with the ultimate goal of creating an effective Human-Robot Collaboration (HRC) system. This system will strategically take advantage of both human flexibility, in terms of adaptability and problem-solving skills, and robot autonomy, in terms of precision and efficiency.

    \captionsetup[figure]{labelsep=period}
	\section{Experiment and Result}
   \begin{figure}[t]
    	\centering
    	\includegraphics[width=0.8\linewidth]{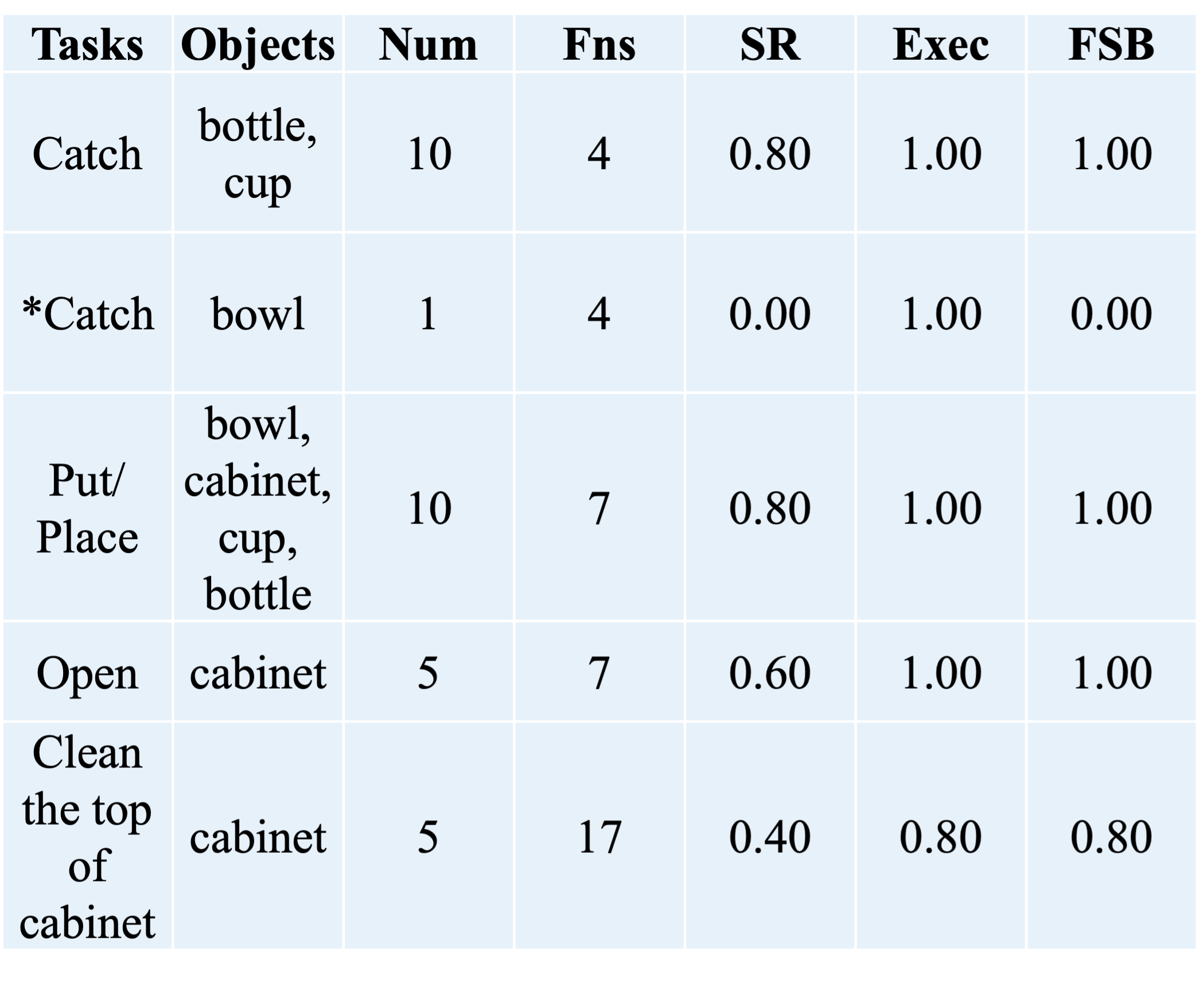}
    	\caption{ Experiment data.  }
    	\label{fig:exp_data}
     \vspace{-8mm}
    \end{figure}
	We conducted multiple experiments by providing "catch", "put", "open" and long-horizon tasks-"clean the top of the cabinet" for several objects to assess their success rates (SR), executability (Exec), and feasibility (FSB). The indicator Num means the number of trials, and Fns shows the motion functions used in completing the task.
     Additionally, Exec is defined as if the task is executable in the environment, and FSB represents if the motion is feasible to reach the goal. 
    The experimental results are presented in Fig. 2. 
    In the case of the indicator "Exec" showing 0.80 in the "clean the top of the cabinet" task, the reasonable explanation is the randomness of LLM, which has a low probability of generating incorrect responses ($\overline{Exec} = 0.2$).
    As for the FBS of 0.00 in the "catch the bowl" task, this outcome can be attributed to the task being impossible to complete due to the default motion function being unsuitable for the target object's shape. In such cases, the DMP-based task correction is used to make necessary demonstrations. 
\
    \section{Conclusion}
    In this work, we have successfully proposed a LLM-based task-planning method. An interface is built to integrate the LLM, perception pipeline, teleoperation system, and DMP-based task correction. The results show that the robot can execute the command from the user with a  considerable success rate for short-horizon tasks like "catch", "put", or "open". Especially, for the task with 0.00 FBS, such as "catch the bowl", DMP-based correction is introduced to improve it. However, for long-horizon tasks, it shows a relatively low success rate. The reason could be the error accumulating with motion. The future work includes the improvement of DMP-based task correction and fine-tuning teleoperation which can complement the error from hardware to improve the success rate and feasibility.

	\section{Acknowledgement}
	This work was supported in part by JST Trilateral AI Research, Japan, under Grant JPMJCR20G8; and in part by JSPS KAKENHI under Grant JP22K14222; and in part by NCGG under Chojuiryou Kenkyukaihatsuhi Nos. 19–5, 21-21.

	\addtolength{\textheight}{-12cm}   
	

 \bibliographystyle{IEEEtran}
 \bibliography{Paper_format_MHS2023/citation}

\end{document}